\documentclass[letterpaper]{article} % DO NOT CHANGE THIS
\usepackage{aaai25}  % DO NOT CHANGE THIS
\usepackage{times}  % DO NOT CHANGE THIS
\usepackage{helvet}  % DO NOT CHANGE THIS
\usepackage{courier}  % DO NOT CHANGE THIS
\usepackage[hyphens]{url}  % DO NOT CHANGE THIS
\usepackage{graphicx} % DO NOT CHANGE THIS
\usepackage{natbib}  % DO NOT CHANGE THIS AND DO NOT ADD ANY OPTIONS TO IT
\usepackage{caption} % DO NOT CHANGE THIS AND DO NOT ADD ANY OPTIONS TO IT
\usepackage{algorithm}
\usepackage{algorithmic}

\usepackage{newfloat}
\usepackage{listings}
\DeclareCaptionStyle{ruled}{labelfont=normalfont,labelsep=colon,strut=off} % DO NOT CHANGE THIS
\floatstyle{ruled}
\newfloat{listing}{tb}{lst}{}
\floatname{listing}{Listing}
\title{Pose as a Modality: A Psychology-Inspired Network for Personality Recognition with a New Multimodal Dataset}
\author {
    Bin Tang\textsuperscript{\rm 1},
    Ke-Qi Pan\textsuperscript{\rm 2},
    Miao Zheng\textsuperscript{\rm 2},
    Ning Zhou\textsuperscript{\rm 2},
    Jia-Lu Sui\textsuperscript{\rm 2},
    Dandan Zhu\textsuperscript{\rm 1},\\
    Cheng-Long Deng\textsuperscript{\rm 2}\footnote{Corresponding author.},
    Shu-Guang Kuai\textsuperscript{\rm 2, \rm3, \rm4}\footnotemark[1]
}
\affiliations {
    \textsuperscript{\rm 1}School of Computer Science and Technology, Shanghai Institute of Artificial Intelligence for Education, Lab of Artificial Intelligence for Education, East China Normal University, Shanghai, China\\
    \textsuperscript{\rm 2}School of Psychology and Cognitive Science, Institute of Brain and Education Innovation, Shanghai Key Laboratory of Mental Health and Psychological Crisis Intervention, Shanghai Key Laboratory of Brain Functional Genomics, Key Laboratory of Brain Functional Genomics, East China Normal University, Shanghai, China\\
    \textsuperscript{\rm 3}Shanghai Center for Brain Science and Brain-Inspired Technology, Shanghai, China\\
    \textsuperscript{\rm 4}NYU-ECNU Institute of Brain and Cognitive Science, New York University Shanghai, Shanghai, China\\
    bintang@stu.ecnu.edu.cn, ddzhu@mail.ecnu.edu.cn,
    \{cldeng, sgkuai\}@psy.ecnu.edu.cn
    
}

\usepackage{bibentry}
\begin{document}

\maketitle

\begin{abstract}
In recent years, predicting Big Five personality traits from multimodal data has received significant attention in artificial intelligence (AI). However, existing computational models often fail to achieve satisfactory performance. Psychological research has shown a strong correlation between pose and personality traits, yet previous research has largely ignored pose data in computational models. To address this gap, we develop a novel multimodal dataset that incorporates full-body pose data. The dataset includes video recordings of 287 participants completing a virtual interview with 36 questions, along with self-reported Big Five personality scores as labels. To effectively utilize this multimodal data, we introduce the Psychology-Inspired Network (PINet), which consists of three key modules: Multimodal Feature Awareness (MFA), Multimodal Feature Interaction (MFI), and Psychology-Informed Modality Correlation Loss (PIMC Loss). The MFA module leverages the Vision Mamba Block to capture comprehensive visual features related to personality, while the MFI module efficiently fuses the multimodal features. The PIMC Loss, grounded in psychological theory, guides the model to emphasize different modalities for different personality dimensions. Experimental results show that the PINet outperforms several state-of-the-art baseline models. Furthermore, the three modules of PINet contribute almost equally to the model’s overall performance. Incorporating pose data significantly enhances the model’s performance, with the pose modality ranking mid-level in importance among the five modalities. These findings address the existing gap in personality-related datasets that lack full-body pose data and provide a new approach for improving the accuracy of personality prediction models, highlighting the importance of integrating psychological insights into AI frameworks.
\end{abstract}


\section{Introduction}
\begin{figure}[t]
\centering
\includegraphics[width=0.8\columnwidth]{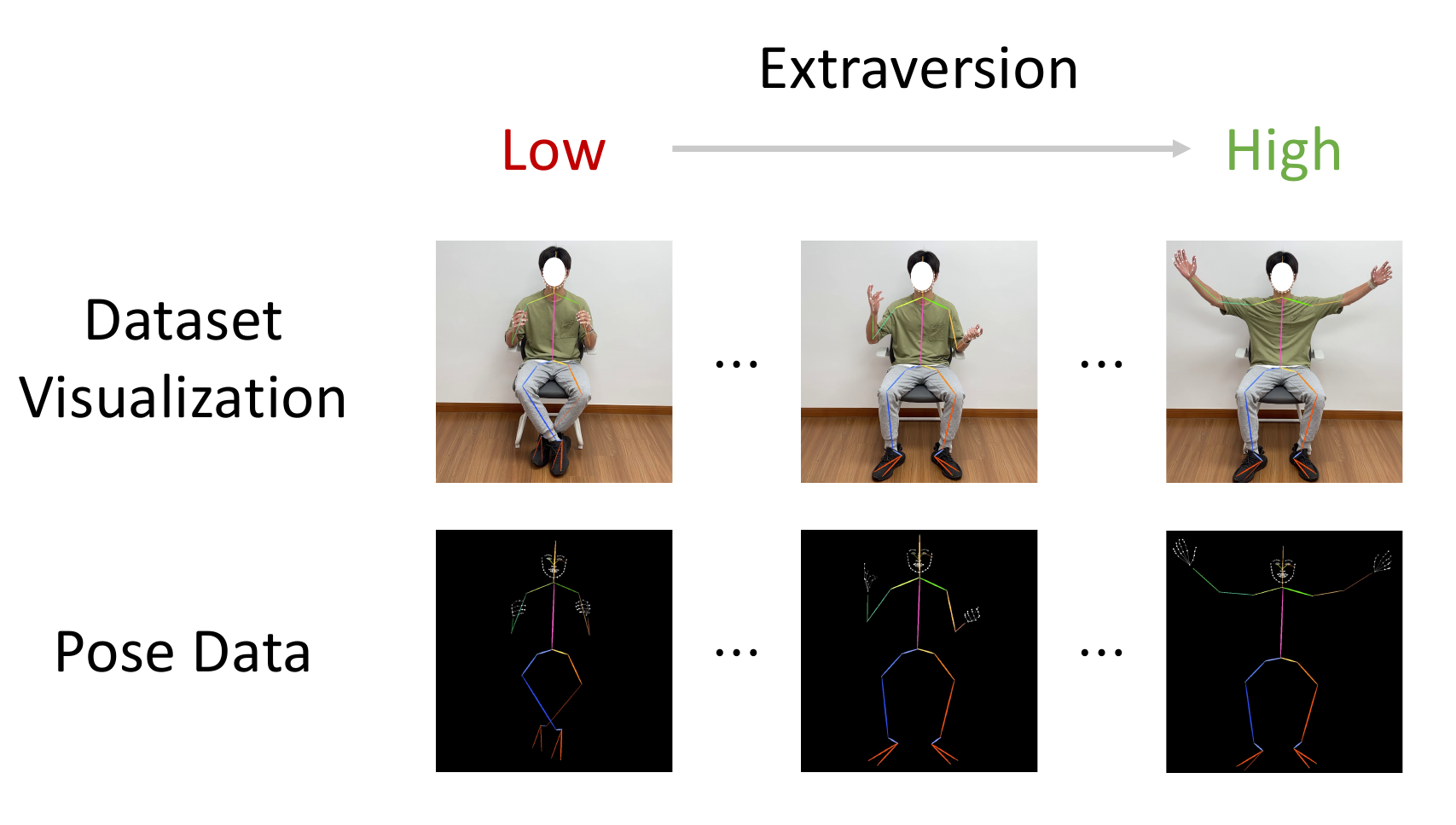}
\caption{Dataset illustration. The first row shows the visualization results using AlphaPose \cite{alphapose}, and the second row shows the dataset's pose data.}
\label{fig:dataset}
\end{figure}

Personality traits are fundamental psychological constructs that influence human behavior and social interactions \cite{sun2022personality}. The Big Five personality model \cite{chen2021big}, which includes five dimensions known as OCEAN (Openness, Conscientiousness, Extraversion, Agreeableness, Neuroticism), is widely recognized and utilized in psychological research. In recent years, the use of machine (deep) learning to analyze multimodal data for detecting the Big Five personality traits has attracted considerable attention from researchers \cite{leekha2024vyaktitvanirdharan, ryumina2024ocean}. Unlike traditional scale-based measures, multimodal personality recognition offers a more effortless assessment process, reduces the likelihood of user manipulation, and enhances efficiency by simultaneously evaluating multiple aspects \cite{hickman2022automated, escalante2020modeling}. As a result, multimodal personality recognition has become increasingly important in fields such as psychology, education, and artificial intelligence \cite{wu2020effects}.

Despite extensive research in computational personality recognition \cite{mehta2020recent}, model accuracy has remained unsatisfactory, making the improvement of predictive performance a key focus for researchers. Psychological research highlights a strong connection between personality traits and pose. Individuals with different personality traits exhibit distinct postural characteristics \cite{kelly2023exploring, thomas2022investigating}. For example, those with higher levels of extraversion tend to use broader and faster gestures, along with increased movements of the head, hands, and legs \cite{conway2023perceiving}. Similarly, varying levels of neuroticism correlate with the relative velocity of pose movements \cite{koppensteiner2013motion}. Given that previous studies have largely overlooked the inclusion of pose data in computational personality models, we hypothesize that incorporating such data could significantly enhance the accuracy of personality predictions. However, two major challenges remain.

The first challenge is the lack of datasets that include frontal full-body pose data. Existing datasets related to personality traits either lack pose data \cite{zafeiriou2017aff}, capture only upper-body pose \cite{escalante2020modeling}, provide side-view full-body pose \cite{giritliouglu2021multimodal}, or collect pose data without the context of speech \cite{miranda2018amigos}. Consequently, no dataset currently available offers high-quality full-body pose data suitable for training models.

The second challenge is that current personality prediction models, such as CRNet \cite{li2020cr} and ResNet \cite{leekha2024vyaktitvanirdharan}, support inputs from text, speech, and facial modalities, but they do not support pose data. Moreover, existing methods predominantly rely on the visual computing paradigm, performing only simple fusion of visual, audio, and text modalities, but do not consider the different contributions of each modality to each dimension of the Big Five personality traits. Psychological research shows that different dimensions of the Big Five personality traits manifest differently across modalities. For example, judgments of Agreeableness primarily rely on facial data, Conscientiousness on body-related data, and Extraversion on full-body data \cite{hu2023first, hu2020integrating}. Therefore, existing models may not achieve good predictions of the Big Five personality traits.

To address these challenges, this study first constructs a multimodal personality recognition dataset that includes whole-body pose data, along with facial, frame, audio, and textual modalities. The dataset comprises full-body videos of 287 participants responding to 36 specific interview questions, annotated with their self-reported Big Five personality scores. We utilize the widely adopted AlphaPose \cite{alphapose} to extract pose information from the visual data (Figure \ref{fig:dataset}). Building on this dataset, we propose the Psychology-Inspired Network (PINet), which consists of three modules: Multimodal Feature Awareness (MFA), Multimodal Feature Interaction (MFI), and Psychology-Informed Modality Correlation Loss (PIMC Loss). The MFA module employs the Vision Mamba Block \cite{zhu2024vision} for the first time to fully capture visual personality features both spatially and temporally. During the MFI process, the Fusion Mamba Block \cite{xie2024fusionmamba} efficiently integrates multimodal information while maintaining linear complexity. The PIMC Loss, grounded in psychological theories, guides the model to emphasize different modalities for different personality dimensions. Notably, PINet serves as an innovative and bidirectional bridge between AI and psychology.

In summary, the contributions of this work are as follows:
\begin{itemize}
    \item We construct a new dataset comprising multimodal data and self-reported Big Five personality scores, addressing the gap in existing personality-related datasets that lack full-body pose data.
    \item We introduce a novel multimodal personality recognition model, the Psychology-Inspired Network (PINet). For the first time, PINet incorporates Mamba technology into personality recognition, significantly enhancing the performance of the visual encoder and multimodal fusion modules.
    \item We propose the innovative Psychology-Informed Modality Correlation Loss (PIMC Loss), which guides the model to emphasize different modalities for various personality dimensions.
    \item Experimental results show that PINet outperforms seven state-of-the-art baseline models. Additionally, integrating pose data significantly improves the performance of various models in predicting personality traits, confirming its crucial role as a modality closely associated with personality traits.

\end{itemize}

\section{Related Work}

\subsection{Personality-related Dataset}
Currently, several personality-related datasets provide multimodal data, but most are limited to common modalities such as speech, text, and facial expressions. For example, the Essays II \cite{tausczik2010psychological} and MyPersonality \cite{stillwell2015mypersonality} datasets focus solely on text, while the First Impressions V2 \cite{escalante2020modeling} and Aff-wild \cite{zafeiriou2017aff} dataset includes face, audio, and text. Although the AMIGOS dataset \cite{miranda2018amigos} includes full-body pose data, it is collected while users are watching videos, which differs significantly from the pose typically exhibited during speech. Therefore, this dataset is not suitable for conventional personality trait modeling and recognition. To date, we have identified only one dataset, SIAP \cite{giritliouglu2021multimodal}, that offers pose data during speech. This dataset records users from frontal, left-rear, and right-front angles. However, the frontal view only provides upper-body images, missing full-body pose data. Although the left-rear and right-front views offer full-body pose data, these side views compromise the accuracy of pose data, which negatively impacts model performance. The study based on the SIAP shows that pose is the least effective modality in predicting the Big Five personality traits among all modalities. In contrast, our findings indicate that pose data has moderate effectiveness among all modalities, underscoring the importance of high-precision pose data.  Additionally, the SIAP dataset is relatively small, containing only 180 video clips. This limited sample size may further affect the model's performance.

\subsection{Multimodal Personality Recognition Methods}
Benefiting from the rapid development of encoders, there are many multimodal personality recognition methods, such as PersEmoN \cite{zhang2019persemon}, SENet \cite{hu2018squeeze}, GatedGCCN \cite{song2022learning}, HRNet \cite{wang2020deep}, EmoFormer \cite{ryumina2024ocean}, ResNet \cite{leekha2024vyaktitvanirdharan}, and CRNet \cite{li2020cr}.

However, these methods either employ a simple CNN or a residual block as the backbone network, or they directly concatenate multimodal features without considering inter-modal correlation. Additionally, previous studies have treated personality recognition merely as a basic classification or regression task, neglecting the internal relationships between the five personality dimensions and multimodal data. With the advent of vision transformers \cite{han2022survey}, the performance of encoders has significantly improved. However, their computational complexity has escalated to quadratic levels. This increased complexity constrains their ability to process more image frames under the same computing power, thereby hindering their capability to capture subtle changes in pose and action. In the multimodal feature fusion module, although multi-head attention \cite{bhojanapalli2020low} considers the interaction between different modalities, it incurs high complexity, leading to more stringent requirements for computing devices.

To this end, we introduce Mamba \cite{gu2023mamba} into the multimodal personality recognition task for the first time and improve both the encoder and the fusion module. Previous studies \cite{zhu2024vision, yang2024vivim} have demonstrated that Mamba can greatly reduce computational complexity. Additionally, we propose the novel PIMC Loss, the first loss function for personality recognition inspired by psychological theory.

\section{Dataset}

\begin{figure}[t]
\centering
\includegraphics[width=0.7\columnwidth]{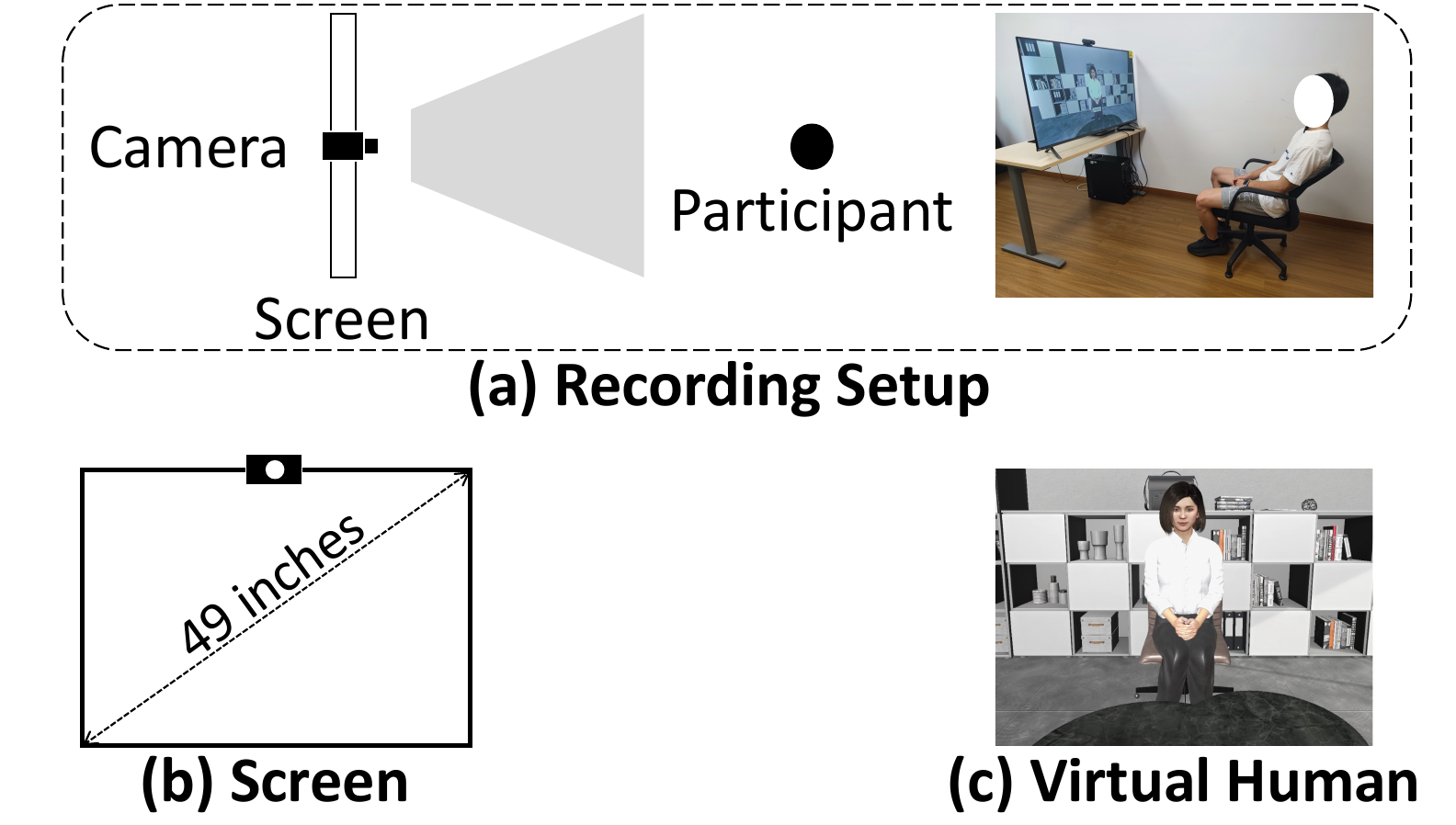}
\caption{The illustration of the data collection scenarios.}
\label{figure10}
\end{figure}

\subsection{Participants}
This dataset includes data from 287 university students (114 males, 173 females; mean age = 21.65 $\pm$ 2.22 years). All participants are native Chinese speakers and have normal or corrected-to-normal visual acuity. They voluntarily take part in the study and sign an informed consent form before the commencement of data collection. The data collection process is approved by the Ethics Committee of the University.

\subsection{Data Collection and Annotation}
Participants engage in a structured interview conducted by a 3D virtual interviewer designed as an Asian woman with a friendly face. Each participant is seated in a room with a 49-inch television around 1.45 meters in front of them. A Logitech BRIO 500 wide-angle camera (Field of View: 90°; Resolution: 1080P; Frame Rate: 30 fps) is mounted on the top edge of the television, facing the participant to capture their full body, as shown in Figure \ref{figure10}.

Participants are required to answer 36 questions adapted from the NEO-FFI-3 \cite{meca2019personality} scale and NEO-PI-R \cite{costa2008revised} scale in sequence from the virtual interviewer. Participants are instructed to limit their response to each question to no more than one minute, with the entire interview lasting 30-60 minutes.

After the interview, all participants complete the NEO-FFI-3 scale, which consists of five dimensions: Openness, Conscientiousness, Extraversion, Agreeableness, and Neuroticism, with 12 questions per dimension, totaling 60 questions. Each question is scored from 0 (strongly disagree) to 4 (strongly agree), with the total score for each dimension ranging from 0 to 48. Following previous research \cite{liao2024open}, we normalize the total score for each dimension to a range of 0 to 1. For more statistics on the dataset, please refer to the supplementary material.

\subsection{Data Preprocessing}
The video data contains homogeneous multimodal data, including visual, audio, and textual information. We use FFmpeg to extract the audio from the videos, FunASR \cite{gao2023funasr} to obtain the participants' spoken text and timestamp information. We then utilize these timestamps to align the three modalities. Subsequently, the multimodal data of 287 participants are divided into 10,332 samples (287 participants $\times$ 36 questions).
 
We use the widely adopted AlphaPose \cite{alphapose} to extract pose information from the visual data. Additionally, MTCNN \cite{zhang2016joint} is used to extract facial data. Finally, the data is organized into five modalities: frame, face, pose, audio, and text.

\section{Methodology}
\begin{figure*}[t]
\centering
\includegraphics[width=1\linewidth]{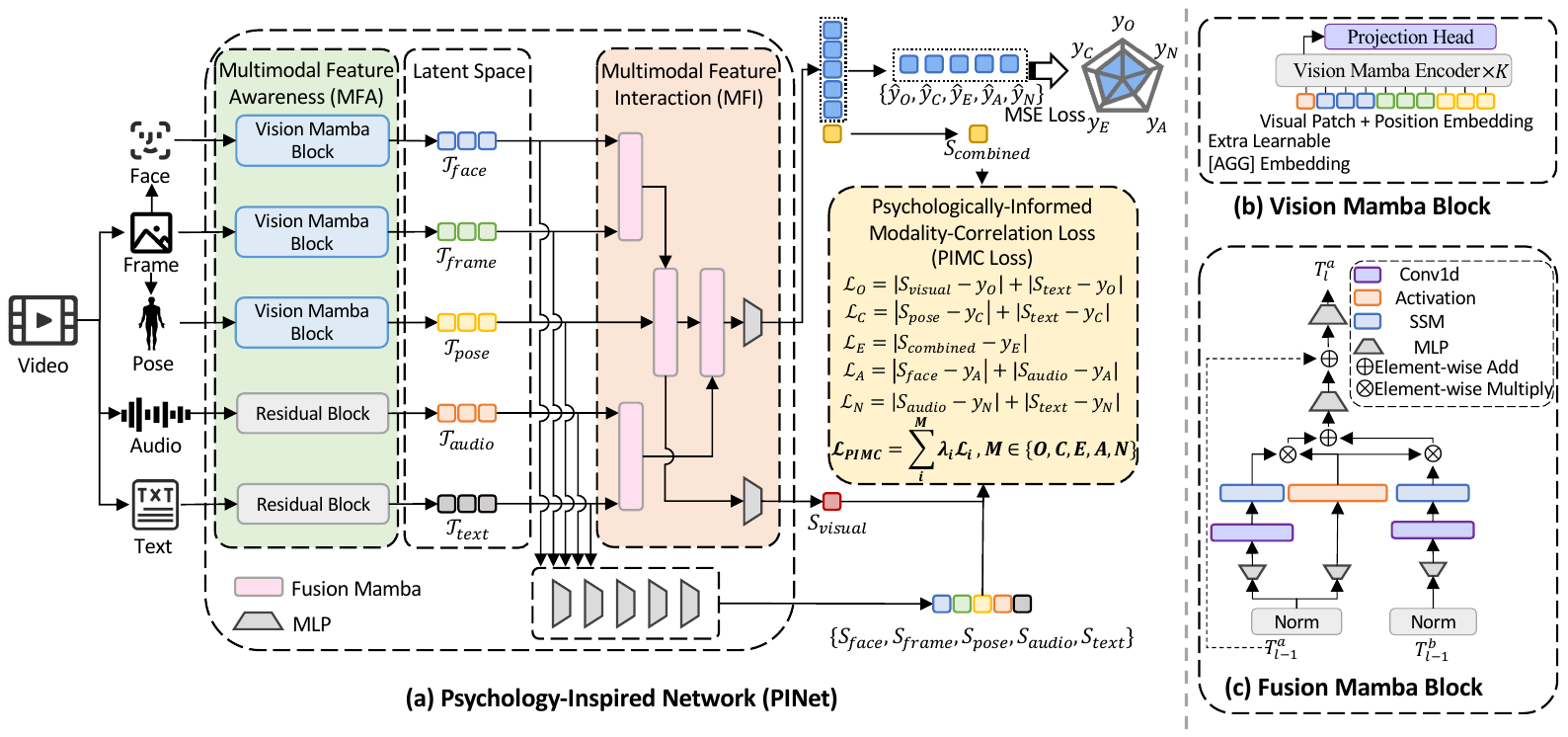}
\caption{(a) The architecture of our PINet. (b) The Vision Mamba Block in MFA. (c) The Fusion Mamba Block in MFI with the symmetrical parts of the Fusion Mamba Block omitted.}

\label{fig:PINet}
\end{figure*}

\subsection{Overview}
Multimodal personality recognition can be regarded as a multimodal multi-task regression problem. In this section, we introduce the details of our proposed PINet, which serves as a novel and bidirectional bridge between AI and psychology. We also explain how PINet processes multimodal multi-task learning. The proposed PINet consists of three modules: Multimodal Feature Awareness (MFA), Multimodal Feature Interaction (MFI), and Psychology-Informed Modality Correlation Loss (PIMC Loss), as illustrated in Figure \ref{fig:PINet}. Both the MFA and MFI modules are based on the advanced state space model (SSM), i.e., Mamba \cite{gu2023mamba}. For details of the SSM, please refer to the supplementary materials.

\subsection{Multimodal Feature Awareness (MFA)}
In the MFA (Encoder) module, there have been many excellent developments, such as the use of residual blocks for feature extraction and the implementation of Vision Transformers to enhance visual feature extraction. Since both audio and text are one-dimensional sequences, their computational load is relatively smaller compared to the three-dimensional sequences of video, and related research has shown that residual blocks perform well. Therefore, this study strictly follows previous researches \cite{liao2024open, li2020cr} in handling audio and text using residual blocks. 

However, the Vision Transformer in the visual modality struggles to capture subtle movements due to its quadratic complexity. The recent Vision Mamba addresses this limitation by extracting visual features with linear complexity. Additionally, the Video Mamba \cite{chen2024video} further extends the Vision Mamba to achieve 3D video understanding. Based on this, we use the Video Mamba as our Vision Mamba Block (VMB), which is shown in Figure \ref{fig:PINet} (b).

Specifically, we first use 3D convolution ($i.e., 1 \times 16 \times 16$) to project the input videos $X_{video} \in R^{T \times H \times W \times 3}$ into $L$ non-overlapping spatiotemporal patches $X^{p} \in R^{L \times D}$, where $X_{video}=\{X_{face}, X_{frame}, X_{pose}\}$,  $X^{p} = \{X^{p}_{face}, X^{p}_{frame}, X^{p}_{pose}\}$, $L=t \times h \times w (t=T, h=\frac{H}{16}, w=\frac{W}{16})$ ($i.e., L$ is the length of sequence), $D$ is the dimension of sequence. The sequence of tokens input to the following VMB is: 
\begin{equation}
X_{visual} = [X_{AGG}, X^{p}] + P_{s} + P_{t},
\end{equation}
where $X_{AGG}$ is the learnable embedding of a special aggregation token whose corresponding output in VMB ($X_{out}^{0}$) is used as the aggregated representation for the entire input sequence. Following previous works\cite{akbari2021vatt}, we add a learnable spatial position embedding $P_{s} \in R^{(hw+1) \times D}$ and the extra temporal one $P_{t} \in R^{n \times D}$ to retain the spatiotemporal position information, since the SSM modeling is sensitive to token position. The tokens $X$ are then passed through by $K$ stacked VMBs, and $X_{AGG}$ token at the final layer is fed into the latent space as visual features ($\mathcal{T}_{visual}$):
%used as the aggregated representation, $T \in R^{L \times D}$, of the entire input sequence and is fed into the latent space:
\begin{equation}
\mathcal{T}_{visual} = VMB(X_{visual}),
\end{equation}
where $VMB(\cdot)$ represents the Vision Mamba Block, $\mathcal{T}_{visual} = \{\mathcal{T}_{face}, \mathcal{T}_{frame}, \mathcal{T}_{pose}\}$. It is worth noting that the reason why these features are denoted by $\mathcal{T}$ is that for Multimodal Feature Interaction (MFI), these features are still treated as tokens. And the   attention features for audio and text modalities are obtained by:
\begin{equation}
\setlength{\arraycolsep}{1pt}
\begin{array}{rcl}
\mathcal{T}_{audio} &= & RB(X_{audio}), \\
\mathcal{T}_{text} &= & RB(X_{text}),
\end{array}
\end{equation}
where $RB(\cdot)$ is the Residual Block, $X_{audio}$ and $X_{text}$ denote the input audios and texts, respectively.
 The multimodal attention features $\mathcal{T} = \{\mathcal{T}_{visual}, \mathcal{T}_{audio}, \mathcal{T}_{text} \}$ is split into two branches, one branch enters the MFI and the other branch is used to obtain the unimodal score $S$, which can be formulated as:
\begin{equation}
S_{i} = MLP(\mathcal{T}_{i}),
\end{equation}
where $i = \{face, frame, pose, audio, text\}$, $MLP(\cdot)$ represents the MLP with sigmoid. The purpose of obtaining the unimodal score $S$ is to calculate the PIMC Loss.

\subsection{Multimodal Feature Interaction (MFI)}
Multimodal feature fusion has always been a challenging problem. The simplest approach is direct concatenation, which, however, overlooks the complementarity and redundancy between modalities. The multi-head attention mechanism addresses this issue effectively but suffers from quadratic computational complexity, leading to low computational efficiency. Fusion mamba \cite{xie2024fusionmamba}, on the other hand, offers a solution to the multi-head attention mechanism’s challenges by employing linear complexity. Initially, it could handle only two modalities. In this study, we expand fusion mamba to suit multimodal personality recognition tasks by integrating four fusion mamba blocks into the Multimodal Feature Interaction module.

Particularly, we project the features (tokens) of the two modalities into a shared space, employing a gating mechanism to encourage complementary feature learning while suppressing redundant features. Additionally, to enhance local features, we incorporate depth-wise convolution within the module, thereby amplifying the encoding capability of local features during the fusion process. The generation of  $\mathcal{T}^a$  and  $\mathcal{T}^b$  follows the process outlined in the Mamba block. Subsequently, we obtain the gating parameter $z$ by projecting  $\mathcal{T}_{l-1}^{'a}$  and use  $z$ to modulate  $\mathcal{T}^a$  and  $\mathcal{T}^b$ . The output sequence $\mathcal{T}_{l}^{a}$ is obtained by residual connection and MLP. The schematic for the Fusion Mamba Block is shown in Figure \ref{fig:PINet} (c). For the pseudocode of the Fusion Mamba Block, please refer to Algorithm 1 in the supplementary material. It is worth noting that Algorithm 1 represents the process of integrating the feature from $\mathcal{T}^b$ into $\mathcal{T}^a$. In other words, a Fusion Mamba consists of two Fusion Mamba Blocks, which are used to integrate one modality into the other, ensuring symmetry between them. The linear layer is designed to further fuse the outputs of the two Fusion Mamba Blocks, which is omitted in Figure \ref{fig:PINet} (a). 
 
The output of PINet $\hat{y}$ can be represented by the following formula:

\begin{equation}
\hat{y} = MLP(MFI(\mathcal{T})),
\end{equation}
where $\hat{y}$ represents the set of predicted values for the five dimensions of the Big Five personality traits $\{\hat{y}_O, \hat{y}_C, \hat{y}_E, \hat{y}_A, \hat{y}_N\}$, $MFI(\cdot)$ represents the Multimodal Feature Interaction.

\subsection{Psychology-Informed Modality Correlation Loss (PIMC Loss)}
Previous studies treat the personality recognition task merely as a regular regression task and ignore the correlations between modalities and personality dimensions. However, psychological studies have shown that different dimensions of the Big Five personality traits may exhibit significant influence in one or more specific modalities rather than uniformly across all modalities \cite{hu2023first, hu2020integrating}.

Therefore, we aim to design a loss function that guides the model to focus on different modalities for different personality dimensions during training, rather than treating all modalities equally. We customize the loss for each Big Five personality dimension based on its definition. Specifically, we hypothesize that openness, which reflects an individual’s acceptance of new ideas, experiences, and cultural diversity, is primarily captured by visual and text modalities. Conscientiousness, related to goal-setting, planning, self-discipline, and execution, is hypothesized to be determined by pose and text. Extroversion, indicating the degree of focus on the external world, sociability, and emotional expression, is believed to be determined by the combined features of the individual. Agreeableness, which describes how an individual interacts with others, being prosocial and cooperative, is felt to be represented by facial expressions and voice. Neuroticism, involving emotional stability and responsiveness, is thought to be mainly determined by pitch and speech content. Based on these assumptions, we design the Psychology-Informed Modality Correlation Loss (PIMC Loss), which includes five loss functions $\mathcal{L}_{O}$, $\mathcal{L}_{C}$,  $\mathcal{L}_{E}$, $\mathcal{L}_{A}$, and $\mathcal{L}_{N}$, for the five dimensions are  simply defined as follows:

\begin{equation}
\setlength{\arraycolsep}{1pt}
\begin{array}{rcl}
\mathcal{L}_{O} &= &MAE(S_{visual}, y_{O}) + MAE(S_{text}, y_{O}),\\
\mathcal{L}_{C} &= &MAE(S_{pose}, y_{C}) + MAE(S_{text}, y_{C}),\\
\mathcal{L}_{E} &= &MAE(S_{combined}, y_{E}),\\
\mathcal{L}_{A} &= &MAE(S_{face}, y_{A}) + MAE(S_{audio}, y_{A}),\\
\mathcal{L}_{N} &= &MAE(S_{audio}, y_{N}) + MAE(S_{text}, y_{N}),
\end{array}
\end{equation}
where $MAE(\cdot)$ represents the mean absolute error, $S_{i}(i = \{face, frame, pose, audio, text\})$ represents the prediction score from unimodal data, $y_{i} (i = \{O, C, E, A, N\}$ represents the ground truth for the five dimensions of the Big Five personality traits. The $S_{visual}$ and $S_{combined}$ represent the prediction scores from the visual modality and all modalities combined, respectively. They are cleverly obtained through the MFI mechanism, as shown in Figure \ref{fig:PINet} (a). The PIMC Loss is defined as:
\begin{equation}
	\mathcal{L}_{PIMC} = \sum^{M}_{i}\lambda_{i}\mathcal{L}_{i}, M \in \{O, C, E, A, N\},
\end{equation}
where $\lambda_{i}$ represents learnable parameters for each modality. The loss function for the whole task is:

\begin{equation}
\mathcal{L}_{MSE} = MSE(\hat{y}, y),
\end{equation}
where $MSE(\cdot)$ represents the mean square error. The final loss function $\mathcal{L}$ is defined as follows:

%The main research objective of this study is a preliminary attempt to combine AI with psychological theory. 
\begin{equation}
\mathcal{L} = \mathcal{L}_{PIMC} + \mathcal{L}_{MSE}.
\end{equation}
Through the design of PIMC Loss, multimodal personality recognition is approached as multi-task learning. Each task is highly correlated and cooperates effectively to enhance the overall model performance, a finding confirmed by the results of ablation experiments with PIMC Loss.

\section{Experiments}

\subsection{Experimental Details}
To fairly compare the performance of different models, all experiments are strictly based on the open-source audio-visual benchmark \cite{liao2024open}, with only the addition of text modality processing. In particular, several models that perform well on the self-reported dataset are used as baselines including CRNet \cite{li2020cr}, ResNet \cite{leekha2024vyaktitvanirdharan}, GatedGCN \cite{song2022learning}, EmoFormer \cite{ryumina2024ocean}, HRNet \cite{wang2020deep}, SENet \cite{hu2018squeeze}, and PersEmoN \cite{zhang2019persemon}. The dataset is divided into training, validation, and test sets in a ratio of approximately 3:1:1 based on participants, with 6,228 samples (173 participants), 2,052 samples (57 participants), and 2,052 samples (57 participants), respectively.  
%The visual tensors (face, frame, and pose) are resized to 224 $\times$ 224 $\times$ 3 and sampled for 60 seconds at 30 fps. 
We use the stochastic gradient descent (SGD) \cite{li2017preconditioned} with an initial learning rate of 0.001, momentum of 0.9, and weight decay of 0.005. All experiments are conducted on the PyTorch platform using six Nvidia A6000 GPUs.

\subsection{Evaluation Metrics}
 
To evaluate the performance of the trained models, we employ four common evaluation metrics: Pearson Correlation Coefficient (PCC), Concordance Correlation Coefficient (CCC), Accuracy (ACC), and Mean Squared Error (MSE) \cite{song2021self, liao2024open}. For PCC, CCC, and ACC, higher values indicate better performance, whereas for MSE, a lower value is preferred.

\subsection{Comparison with Baselines}

\begin{table}[ht]
\centering
{\fontsize{9pt}{9pt}\selectfont
\begin{tabular}{lcccc}
\hline
Model           & PCC $\uparrow$   & CCC $\uparrow$   & ACC $\uparrow$   & MSE $\downarrow$   \\ \hline
PersEmoN        & 0.0359 & -0.0212 & 0.8575 & 0.0320 \\
SENet           & 0.0567 & 0.0549 & 0.8628 & 0.0297 \\
GatedGCN		& 0.0777 & 0.0345 & \underline{0.8651} & 0.0289 \\
HRNet           & 0.0924 & 0.0284 & 0.8557 & 0.0327 \\
EmoFormer		& 0.1046 & 0.0320 & 0.8640 & \underline{0.0293} \\
ResNet          & 0.1054 & 0.0542 & 0.8506 & 0.0354 \\
CRNet           & \underline{0.1141} & \underline{0.1124} & 0.8577 & 0.0321 \\

\textbf{Our PINet}      & \textbf{0.2509} & \textbf{0.2706} & \textbf{0.8679} & \textbf{0.0277} \\ \hline
\end{tabular}
}
\caption{Performance comparison of our proposed PINet with baselines. The \textbf{bold} result indicates the best performance, while the \underline{underlined} result indicates the second-best performance.}
\label{tab:baselines}
\end{table}

\begin{table*}[ht]
\setlength{\tabcolsep}{1mm}
\centering
{\fontsize{9pt}{9pt}\selectfont
\begin{tabular}{@{}lccccc|ccccc@{}}
\hline
& \multicolumn{5}{c|}{Pearson Correlation Coefficient (PCC) $\uparrow$} & \multicolumn{5}{c}{Concordance Correlation Coefficient (CCC) $\uparrow$} \\
\raisebox{1ex}[0pt]{Model} & Open & Consc & Extrav & Agree & Neuro & Open & Consc & Extrav & Agree & Neuro  \\
\hline
PersEmoN    & 0.0333 & 0.0508 & 0.0192 & 0.0448 & 0.0316  & 0.0002 & 0.0431 & -0.1732 & 0.0344 & -0.0103  \\
SENet       & 0.0306 & 0.0019 & 0.1176 & 0.0468 & 0.0864  & 0.0296 & 0.0018 & 0.1139  & 0.0453 & 0.0837    \\
GatedGCN    & 0.0403	 & 0.0818 & 0.1183 & 0.0579 & 0.0903 & 0.0197 & 0.0184	 & 0.0681	 & 0.0393 & 0.0267\\
HRNet       & 0.0619 & 0.0885 & 0.1193 & 0.1323 & 0.0599  & 0.0190  & 0.0272 & 0.0367  & 0.0407 & 0.0184    \\
EmoFormer  &0.0584	&0.1139	&0.1097	&0.1487	&0.0921 &0.0179	&0.0359	&0.0387	&0.0491	&0.0183\\
ResNet      & 0.0288 & \underline{0.1159} & 0.1164 & \underline{0.1522} & \underline{0.1138}  & 0.0148 & 0.0596 & 0.0598  & 0.0783 & 0.0585    \\
CRNet       & \underline{0.1258} & 0.0885 & \underline{0.1334} & 0.1512 & 0.0747  & \underline{0.1239} & \underline{0.0842} & \underline{0.1314}  & \underline{0.1489} & \underline{0.0736}    \\
\textbf{Our PINet}  & \textbf{0.3335} & \textbf{0.1463} & \textbf{0.3146} & \textbf{0.2788} & \textbf{0.1812}  & \textbf{0.3447} & \textbf{0.1907} & \textbf{0.3184}  & \textbf{0.2821} & \textbf{0.2172}    \\
\hline
\end{tabular}
}
\caption{PCC and CCC values of our PINet and baseline models for predicting each dimension of the personality traits.}
\label{tab:pcc-ccc}
\end{table*}

\begin{table*}[ht]
\centering
{\fontsize{9pt}{10pt}\selectfont
\begin{tabular}{@{}lccccc|ccccc@{}}
\hline
& \multicolumn{5}{c|}{Accuracy (ACC) $\uparrow$} & \multicolumn{5}{c}{Mean Squared Error (MSE) $\downarrow$} \\
\raisebox{1ex}[0pt]{Model} & Open & Consc & Extrav & Agree & Neuro & Open & Consc & Extrav & Agree & Neuro  \\
\hline
PersEmoN    & 0.8526 & 0.8482 & 0.8790  & 0.8515 & 0.8562  & 0.0339 & 0.0360  & 0.0227 & 0.0345 & 0.0328  \\
SENet       & 0.8557 & 0.8630  & 0.8789 & 0.8541 & 0.8623  & 0.0328 & 0.0301 & 0.0230  & 0.0334 & 0.0294  \\
GatedGCN   &\underline{0.8591}	&\underline{0.8692}	&0.8793	&0.8535	&\underline{0.8644} & 0.0310	&0.0304	&0.0219	&\underline{0.0323}	&\underline{0.0291} \\
HRNet       & 0.8514 & 0.8531 & 0.8771 & 0.8430  & 0.8538  & 0.0341 & 0.0343 & 0.0233 & 0.0388 & 0.0331  \\
EmoFormer  &0.8535	&0.8646	&\underline{0.8824}	&0.8563	&0.8631 &0.0313	&\underline{0.0292}	&\underline{0.0216}	&0.0340	&0.0305\\
ResNet      & 0.8440  & 0.8438 & 0.8722 & 0.8409 & 0.8521  & 0.0383 & 0.0379 & 0.0261 & 0.0396 & 0.0354  \\
CRNet       & 0.8525 & 0.8505 & 0.8773 & \textbf{0.8578} & 0.8503  & 0.0340  & 0.0355 & 0.0240  & 0.0324 & 0.0347  \\
\textbf{Our PINet}  & \textbf{0.8622} & \textbf{0.8694} & \textbf{0.8857} & \underline{0.8572} & \textbf{0.8649}  & \textbf{0.0303} & \textbf{0.0267} & \textbf{0.0209} & \textbf{0.0321} & \textbf{0.0283}  \\
\hline
\end{tabular}
}
\caption{ACC and MSE values of our PINet and baseline models for predicting each dimension of the personality traits.}
\label{tab:acc-mse}
\end{table*}

Table \ref{tab:baselines}, Table \ref{tab:pcc-ccc}, and Table \ref{tab:acc-mse} present the performance comparison between PINet and other methods on our dataset. Table \ref{tab:baselines} shows the average results across the five personality trait dimensions, while Table \ref{tab:pcc-ccc} and Table \ref{tab:acc-mse} provide detailed results for each dimension of the traits. 

The average results across the five dimensions demonstrate that our PINet model significantly outperforms other baseline models across all four evaluation metrics. Notably, the PCC and CCC scores of PINet are higher than those of the second-best model, CRNet, by 0.1368 and 0.1508, respectively, representing improvements of 119.9\% and 134.2\% (Table \ref{tab:baselines}). These findings are further supported by the detailed results for each dimension of the traits. Among all models, PINet achieves the highest PCC and CCC scores across all five dimensions (Table \ref{tab:pcc-ccc}), and it also achieves the best performance in ACC for four dimensions (Open, Consc, Extrav, Neuro) and in MSE for five dimensions (Table \ref{tab:acc-mse}). These results indicate that, our proposed PINet model significantly improves the accuracy of predicting the traits. For the experimental results of PINet and the baselines on the public dataset, please refer to the supplementary material.

\subsection{Ablation Study}
\begin{figure}[t]
\centering
\includegraphics[width=0.94\columnwidth]{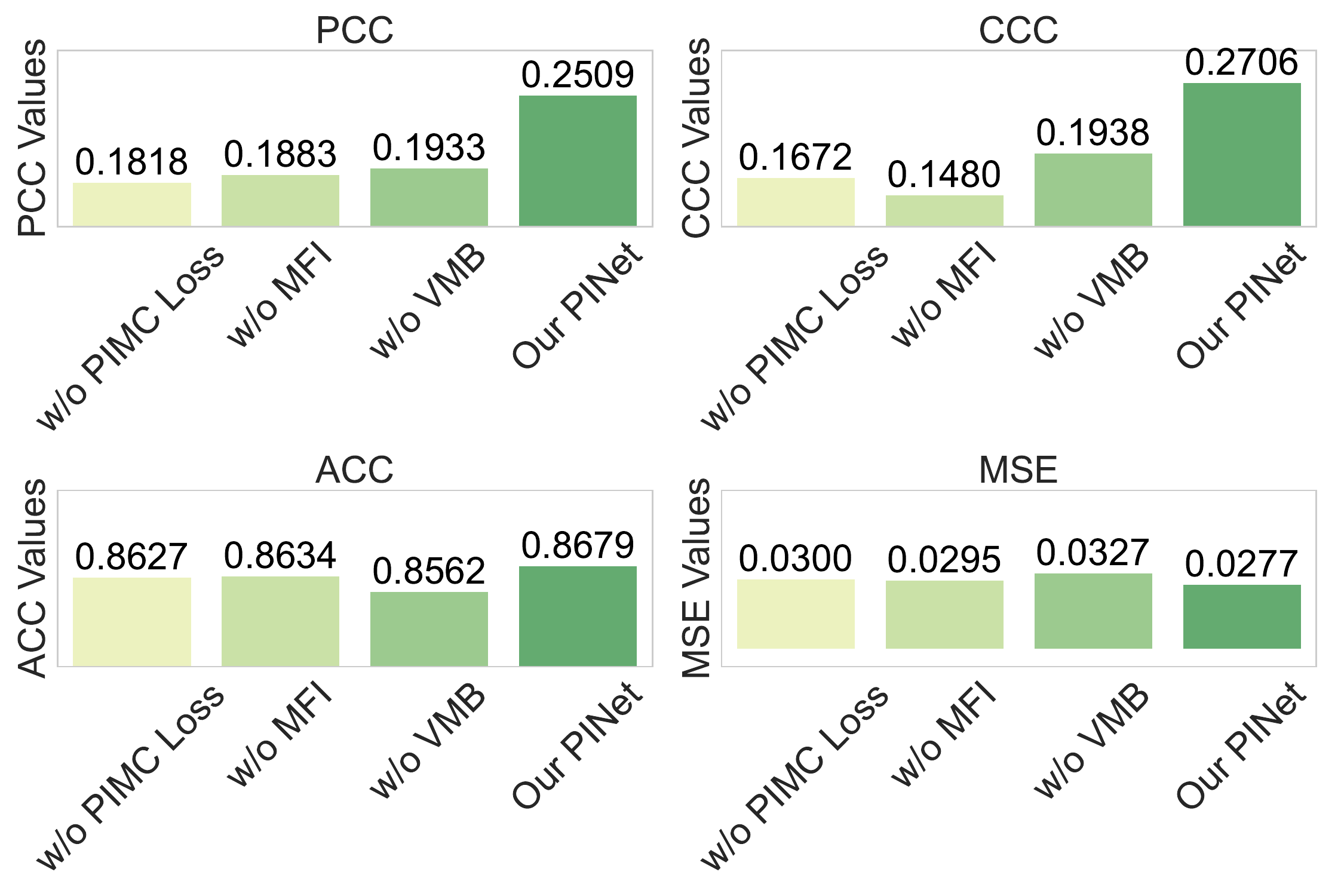}
\caption{Ablation study with PIMC Loss, MFI and VMB.}
\label{figure4}
\end{figure}

\begin{figure}[t]
\centering
\includegraphics[width=0.94\columnwidth]{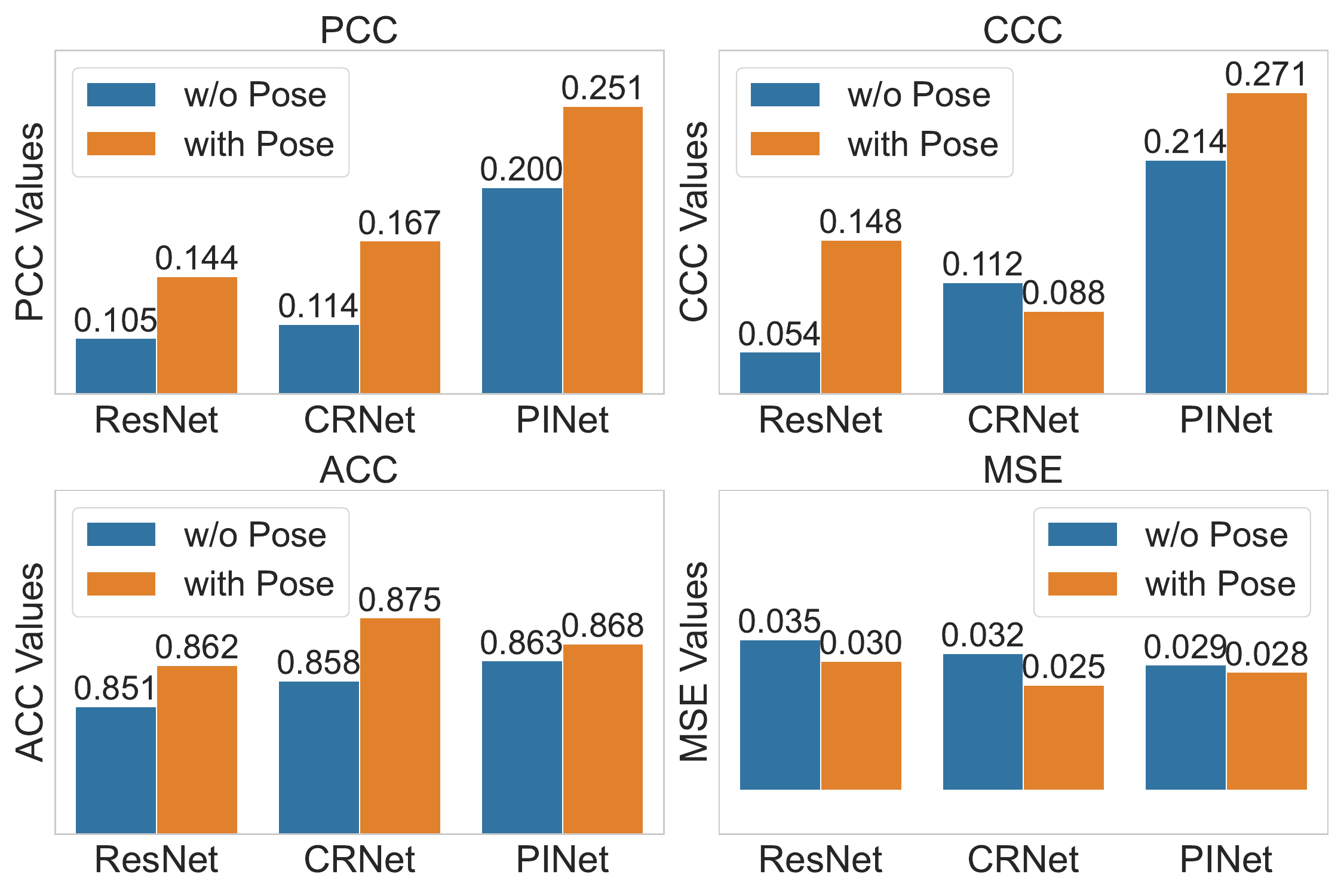}
\caption{Ablation study with pose data.}
\label{fig:Ablation pose}
\end{figure}

We further conduct ablation studies to assess the importance of PINet’s three modules. To evaluate the significance of the VMB module, we replace it with a conventional Vision Transformer (w/o VMB) while keeping the other modules of PINet unchanged. Similarly, to assess the MFI module, we replace it with a simple concatenation method (w/o MFI). For evaluating the PIMC Loss module, we directly remove it from the model (w/o PIMC Loss). The average results across the five dimensions of the traits (as shown in Figure \ref{figure4}) show that, compared to the PINet, removing any of the three modules leads to a significant drop in PCC and CCC values, a slight decrease in ACC, and an increase in MSE. Moreover, the values of the three modules are quite close in PCC, ACC, and MSE. The results for each dimension are similar to the average results across all dimensions (please refer to supplementary Table 1 and Table 2). These findings demonstrate that all three modules are crucial for improving the prediction of the traits and that their contributions are comparable. 

In addition, we conduct another ablation study to evaluate the contribution of pose data to prediction accuracy using the common ResNet, CRNet, and our PINet. We compare the prediction results of models with and without pose data for the traits. The average results across the five dimensions are shown in Figure \ref{fig:Ablation pose}. For all three models, those with pose data significantly outperform those without pose data in PCC, CCC, ACC, and MSE, except for the CCC for CRNet. Notably, in terms of PCC, the values of the PINet, CRNet, and ResNet models with pose data increase by 36.8\%, 45.9\%, and 25.5\%, respectively, compared to models without pose data. For CCC, the values of the PINet and ResNet models increase by 172.5\% and 26.3\%, respectively. These findings are further supported by the results for each dimension of the  traits (please refer to supplementary Table 3 and Table 4). These results demonstrate the importance of pose data in improving the prediction of the traits.

\begin{figure}[t]
\centering
\includegraphics[width=0.7\columnwidth]{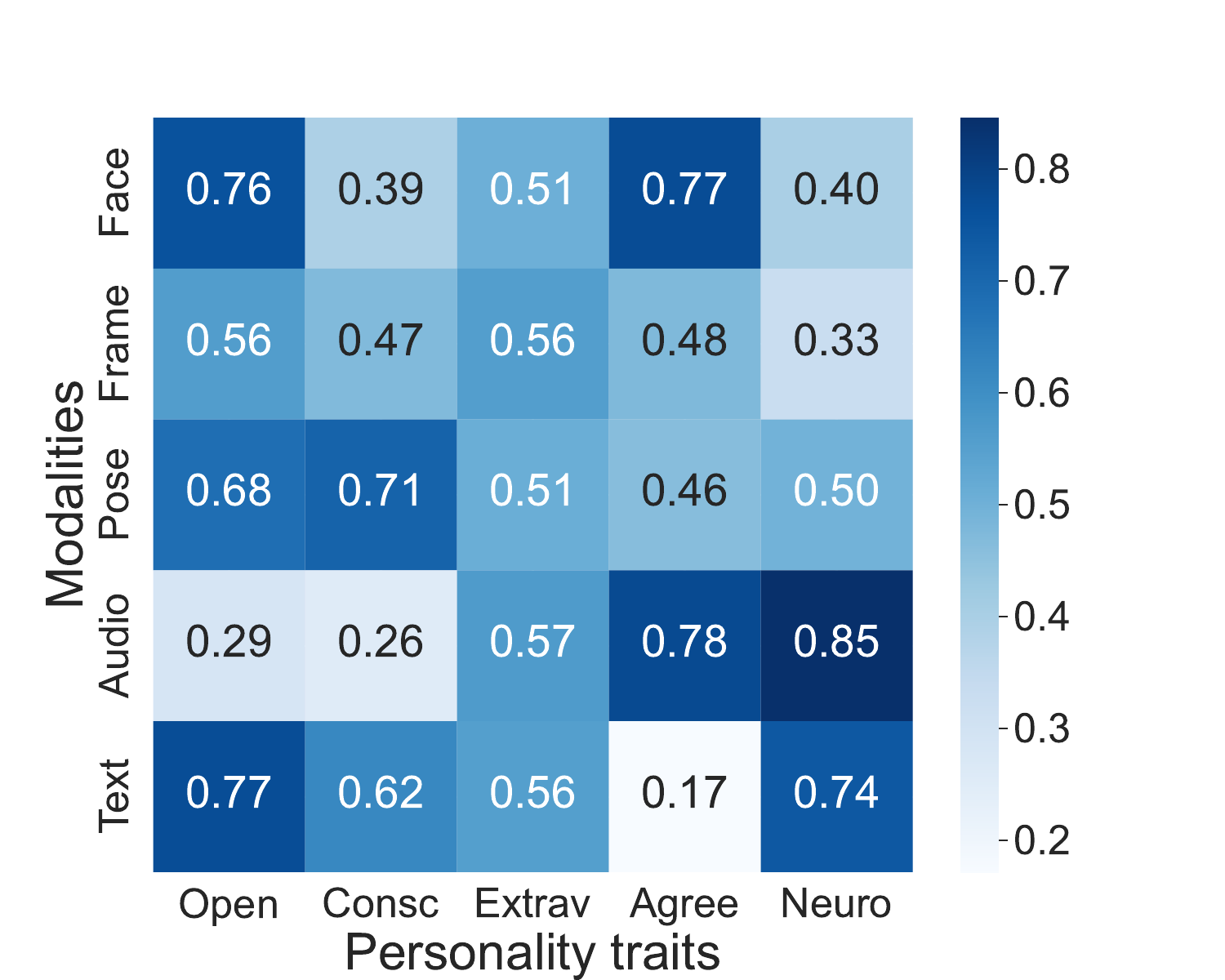}
\caption{The relative PCC values of the single modality to each personality trait.}
\label{fig:Correlation}
\end{figure}

Furthermore, we examine the relative importance of the pose modality compared to other modalities, as shown in Figure \ref{fig:Correlation}. We train the revised PINet model, without the MFI and PIMC Loss modules, using only a single modality (face, frame, pose, audio, text) at a time and calculated the PCC for each modality model. We then calculate the relative PCC by dividing them by the PCC value obtained when training PINet with all modalities, yielding values between 0 and 1. The results show that pose's average relative PCC ranked third, higher than face and frame but lower than text and audio. For each dimension of the traits, the relative PCC of pose ranks third for Open, first for Consc, tied for fourth for Extra, fourth for Agree, and third for Neuro. These findings suggest that the importance of pose is moderate among all modalities.

\section{Conclusion and Future Work}
In this study, we first create a dataset consisting of frontal full-body video recordings of 287 participants during interviews, along with self-reported Big Five personality scores as labels. This dataset addresses the gap in existing personality-related datasets that need more high-quality whole-body pose data. Second, to better utilize multimodal data such as pose, we propose a new network called PINet, which consists of three modules: MFI, VMB, and PIMC Loss, based on findings from psychological research. The PINet network assigns different weights to different modalities when predicting each dimension of the Big Five personality traits. Experimental results demonstrate that our proposed PINet outperforms other state-of-the-art baseline models, and the three modules of PINet contribute similarly to the overall performance, which indicates that the psychology-based PINet model can significantly enhance the accuracy of Big Five personality trait predictions. Additionally, the pose modality significantly improves the accuracy of Big Five personality trait predictions. Among the five modalities (face, frame, pose, audio, text), the importance of the pose modality ranks in the middle.

In future research, we plan to further expand the dataset by collecting observer-reported Big Five personality scores to verify the importance of the pose modality and investigate the differences in model prediction between self-reported and observer-reported personality traits. Additionally, we will test our PINet model with the observer-reported label to validate the generalizability of our PINet.

\section{Acknowledgments}
This work is supported by the National Science and Technology Innovation 2030 Major Program [grant number 2022ZD0205103], the National Natural Science Foundation of China [grant number 32100879], and the Fundamental Research Funds for the Central Universities.

\bibliography{aaai25}

\end{document}